\pdfoutput=1

\documentclass[11pt]{article}

\usepackage[]{EACL2023}
\usepackage{times}
\usepackage{latexsym}
\usepackage{graphicx}
\usepackage{float}
\usepackage{subfig}
\usepackage{array}
\usepackage{multirow}
\usepackage[normalem]{ulem}
\usepackage{wrapfig}
 \usepackage{booktabs}
\usepackage[T1]{fontenc}

\usepackage[utf8]{inputenc}

\usepackage{microtype}

\usepackage{inconsolata}
\usepackage[prependcaption,textsize=tiny]{todonotes}

\newcommand\crule[3][black]{\textcolor{#1}{\rule{#2}{#3}}}
\definecolor{lightg}{rgb}{0.8509803922,	0.9215686275,	0.831372549}
\definecolor{lorg}{rgb}{0.9882352941,	0.8980392157,	0.8117647059}
\definecolor{lpink}{rgb}{0.9215686275,	0.8156862745,	0.862745098}
\definecolor{lvio}{rgb}{0.8509803922,	0.8196078431,	0.9098039216}
\definecolor{special_red}{rgb}{0.65098039215,0.30196078431,0.47450980392}

\newcommand*{\affmark}[1][*]{\textsuperscript{#1}}

\title{Systematic Investigation of Strategies Tailored for Low-Resource Settings for Low-Resource Dependency Parsing}

\author{Jivnesh Sandhan\affmark[1], \textbf{Laxmidhar Behera\affmark[1,3] and Pawan Goyal\affmark[2]}\\
\affmark[1]IIT Kanpur, \affmark[2]IIT Kharagpur, \affmark[3]IIT Mandi\\
\texttt{jivnesh@iitk.ac.in,pawang@cse.iitkgp.ac.in}}

\begin{document}
\maketitle
\begin{abstract}
 In this work, we focus on low-resource dependency parsing for multiple languages. Several strategies are tailored to enhance performance in low-resource scenarios.
 While these are well-known to the community, it is not trivial to select the best-performing combination of these strategies for a low-resource language that we are interested in, and not much attention has been given to measuring the efficacy of these strategies.
 We experiment with 5 low-resource strategies for our ensembled approach on 7 Universal Dependency (UD) low-resource languages.
Our exhaustive experimentation on these languages supports the effective improvements for languages not covered in pretrained models.
We show a successful application of the ensembled system on a truly low-resource language Sanskrit.\footnote{The code and data are available at: \url{https://github.com/Jivnesh/SanDP}.}
\end{abstract}

\section{Introduction}
Recently, the supervised learning paradigm has dramatically increased the state-of-the-art performance for the dependency parsing task for resource-rich languages~\cite{chen-manning-2014-fast,dyer-etal-2015-transition,kiperwasser-goldberg-2016-simple,DBLP:conf/iclr/DozatM17,kulmizev-etal-2019-deep}.
However, only a handful of resource-rich languages are able to take advantage, and many low-resource languages are far from these benefits \cite{joshi-etal-2020-state,more19,zeman-etal-2018-conll}. 

  In literature, several strategies have been proposed to enhance performance in low-resource scenarios, such as data augmentation \cite{sahin-steedman-2018-data,gulordava-etal-2018-colorless}, cross/mono-lingual pretraining \cite{conneau-etal-2020-unsupervised,peters-etal-2018-deep,kondratyuk-straka-2019-75}, sequential transfer learning \cite{ruder-etal-2019-transfer}, multi-task learning \cite{nguyen-verspoor-2018-improved}, cross-lingual transfer~\cite{cross_survey} and self-training \cite{rotman2019deep,clark-etal-2018-semi}.
    However, not much attention has been given to measuring the efficacy of the existing low-resource strategies well-known to the community for low-resource dependency parsing \cite{vania-etal-2019-systematic}. 
  This is essential to assess their utility for low-resource languages \cite{hedderich-etal-2021-survey} before inventing novel ways to tackle data sparsity.

In this work, we systematically explore 5 pragmatic strategies for low-resource settings on 7 languages. We experiment with low-resource strategies such as data augmentation, sequential transfer learning, cross/mono-lingual pretraining, multi-task learning and self-training. We investigate: (1) How is the trend in performance of each strategy across various languages? Whether the choice of best performing variant of each strategy is language dependent? (2) We integrate the best performing variant of each strategy and call the resulting system as the ensembled system. Do all the strategies contribute towards performance gain in the ensembled system? How well does this ensemble approach generalize across multiple low-resource languages?  (3) 
How far can we push a purely data-driven ensemble system using the best-performing low-resource strategies? Can this simple ensemble approach outperform the state-of-the-art of a low-resource language? We argue that while it may sound like a simple application of techniques well known to the community; it is non-trivial to select the best performing combination for a target low-resource language.

Our exhaustive experimentation empirically establishes the effective generalization ability of the ensembled system on 7 languages and shows average absolute gains of 5.2/6.2 points Unlabelled/Labelled Attachment Score (UAS/LAS) over strong baseline \cite{dozat2017stanford}. 
Notably, our ensembled system shows substantial improvements for the languages not covered in pretrained models.
Finally, we show a successful application of the ensembled system on a truly low-resource language Sanskrit.
We find that the ensembled system outperforms the state-of-the-art system \cite{krishna-etal-2020-keep} for Sanskrit by 1.2 points absolute gain in terms of UAS and shows comparable performance in terms of LAS (\S~\ref{experiments}).

\section{Investigation of Strategies Tailored for Low-resource Settings}
\label{5_strategies}
 We explore 5 strategies specially tailored for low-resource settings on 7 languages, and integrate the best performing strategy of each category in our ensembled system (Table~\ref{table:multilingual_results}). We utilize \newcite{DBLP:conf/iclr/DozatM17} as a base system for all the experiments, henceforth referred to as BiAFF.

\paragraph{Language selection criteria:} 
 We choose low-resource languages with less than 2,500 training samples from 4 different typological families such that each language belongs to a unique sub-family.   In order to accommodate a low-resource tailored pretraining \cite{sandhan-etal-2021-little}, we choose languages that have explicit morphological information. Additionally, we divide the set of languages into the languages covered/not-covered in the multilingual language model’s pretraining: (1) Covered: Arabic (ar), Greek (el), Hungarian (hu) (2) Not covered:  Wolof (wo), Gothic (got), Coptic (cop) and Sanskrit (san).

\begin{figure}[!h]
\centering
{\includegraphics[width=1\linewidth]{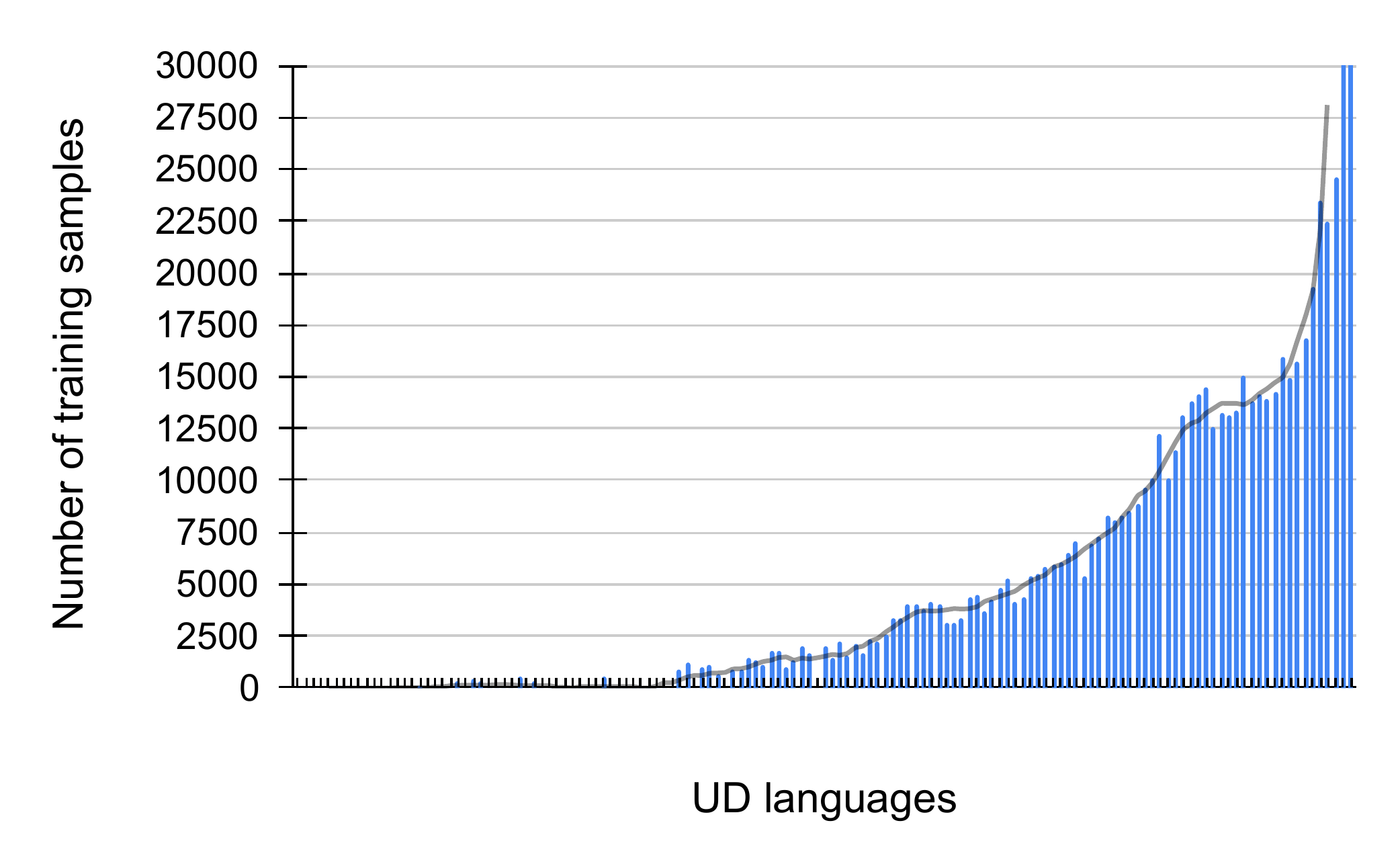}}
\caption{Plot for number of training samples vs. UD languages available in UD-2.6.}
\label{fig:UD_vs_train}
\end{figure}
\noindent Figure~\ref{fig:UD_vs_train} illustrates the number of training samples available for all UD languages in UD-2.6.
  \newcite{hedderich-etal-2021-survey} ask ``How low is low-resource?'' and suggest that the threshold of low-resource is task and language-dependent. The low-resource settings can be seen as a continuum of resource availability due to the absence of a hard threshold. More efforts should focus on evaluating low-resource strategies across multiple languages for a fair comparison between these strategies. Therefore, we select a threshold for the languages with less than 2,500 training samples (Figure~\ref{fig:UD_vs_train}). 
   We restrict ourselves to the setting where the target low-resource language does not have a high-resource related language that could possibly facilitate the positive cross-lingual or zero-shot transfer 
   \cite{vulic-etal-2019-really,pires-etal-2019-multilingual,sogaard-etal-2018-limitations,de-lhoneux-etal-2018-parameter,smith-etal-2018-82}. Thus, we do not consider low-resource languages with only a test set available. Also, we do not consider cross-lingual transfer~\cite{duong2015low,ahmad-etal-2019-cross,vania-etal-2019-systematic,cross_survey} strategy in our study.

\paragraph{Dataset and metric:} For each of these  7 low-resource languages from Universal Dependencies (UD-2.6) \cite{10.1162/coli_a_00402},
  following \newcite{rotman2019deep}, we use 500 data points for training (allocating equal power to each language for fair comparison) and the original dev/test split as dev/test set. Additionally, 1000 morphologically tagged data points (without dependency annotations) are used for self-training and pretraining.
   We use sentence level macro averaged UAS/LAS metric for evaluation.

\paragraph{Hyper-parameters:}
\label{hyper-parameter}
  For SeqTraL variants, we use the exact same encoder as \newcite{ma-etal-2018-stack} with 2 Bi-LSTM layers and decoder with fully connected layer followed by softmax layer. For the ensembled system, we adopt BiAFF's codebase by \newcite{ma-etal-2018-stack} with the hyper-parameters setting as follows: the batch size as 16, training iterations as 100, a dropout rate as 0.33, the number of stacked Bi-LSTM layers as 2, learning rate as 0.002 and the remaining parameters as the same as \newcite{ma-etal-2018-stack}. We release our codebase publicly under creative-common licence. 
 \paragraph{Computing Infrastructure Used:} We primarily use RTX-2080, 12 GB GPU memory, 4352 GPU Cores computing infrastructure for our experiments.

 \begin{table*}[bht]
\begin{small}
    \centering
\resizebox{1\textwidth}{!}{
 \begin{tabular}{ccccccccccccccccc}
\toprule
& &\multicolumn{2}{c}{el} &\multicolumn{2}{c}{ar} &\multicolumn{2}{c}{hu} &\multicolumn{2}{c}{got}&\multicolumn{2}{c}{cop}
  &\multicolumn{2}{c}{wo}&\multicolumn{2}{c}{san}
 \\\cmidrule(r){2-2}\cmidrule(r){3-4}\cmidrule(l){5-6}\cmidrule(l){7-8}\cmidrule(l){9-10}\cmidrule(l){11-12} \cmidrule(l){13-14} \cmidrule(l){15-16}
Strategy&Model &UAS&LAS    & UAS &LAS   & UAS &LAS & UAS &LAS & UAS &LAS &UAS&LAS  &UAS&LAS  
\\\cmidrule(r){2-2}\cmidrule(r){3-4}\cmidrule(l){5-6}\cmidrule(l){7-8}\cmidrule(l){9-10}\cmidrule(l){11-12} \cmidrule(l){13-14} \cmidrule(l){15-16}
&BiAFF      & 86.61          & 82.23          & \textbf{79.89} & \textbf{73.13} & \textbf{80.51} & \textbf{75.11} & 75.66          & \textbf{69.42} & 87.33          & \textbf{84.48} & \textbf{82.63} & \textbf{77.83} & 75.47          & 65.77          \\
\cmidrule(r){2-2}\cmidrule(r){3-4}\cmidrule(l){5-6}\cmidrule(l){7-8}\cmidrule(l){9-10}\cmidrule(l){11-12} \cmidrule(l){13-14} \cmidrule(l){15-16}

&Cropping &      85.98          & 81.76          & 79.42          & 73.03          & 77.46          & 71.62          & 75.72          & 68.25          & 86.05          & 82.95          & 80.38          & 75.39          & 73.82          & 63.46          \\
Data aug.&Rotation &      86.39          & 82.19          & 79.22          & 72.79          & 77.33          & 71.16          & 76.19          & 68.89          & 86.27          & 83.27          & 80.66          & 76.05          & 75.58          & 65.14          \\
&Nonce   &    \textbf{87.43} & \textbf{82.60} & 79.52          & 73.00          & 79.56          & 69.50          & \textbf{76.42} & 67.74          & \textbf{87.64} & 83.52          & 81.97          & 76.37          & \textbf{77.25} & \textbf{66.30} \\
\cmidrule(r){2-2}\cmidrule(r){3-4}\cmidrule(l){5-6}\cmidrule(l){7-8}\cmidrule(l){9-10}\cmidrule(l){11-12} \cmidrule(l){13-14} \cmidrule(l){15-16}
&BiAFF+mBERT &     91.41          & 87.89          & 83.50          & 76.30          & 85.20          & 77.50          & 64.20          & 53.20          & 33.70          & 15.60          & 71.50          & 61.40          & 71.40          & 55.12          \\
Pretraining&BiAFF+XLM-R &  \textbf{93.61} & \textbf{90.85} & \textbf{86.04} & \textbf{79.55} & \textbf{89.05} & \textbf{83.80} & -              & -              & -              & -              & -              & -              & 78.43          & 66.72          \\
&BiAFF+LCM  &     89.00          & 85.83          & 82.49          & 76.67          & 83.22          & 78.49          & \textbf{79.88} & \textbf{74.65} & \textbf{88.79} & \textbf{86.02} & \textbf{85.85} & \textbf{81.66} & \textbf{81.63} & \textbf{73.86} \\
\cmidrule(r){2-2}\cmidrule(r){3-4}\cmidrule(l){5-6}\cmidrule(l){7-8}\cmidrule(l){9-10}\cmidrule(l){11-12} \cmidrule(l){13-14} \cmidrule(l){15-16}
&SelfTrain  &    86.78          & 82.25          & 80.86          & 74.45          & 80.62          & 75.09          & 76.96          & 70.15          & 87.95          & 85.33          & 83.83          & 78.80          & 77.53          & 66.59          \\
Self-training&CVT  &     80.53          & 77.37          & 76.21          & 71.87          & 75.21          & 70.01          & 69.43          & 63.59          & 79.32          & 74.21          & 73.21          & 69.50          & 69.21          & 56.21          \\
&DCST    &   \textbf{88.26} & \textbf{84.09} & \textbf{82.21} & \textbf{75.78} & \textbf{82.85} & \textbf{77.65} & \textbf{79.52} & \textbf{72.91} & \textbf{88.85} & \textbf{85.53} & \textbf{85.51} & \textbf{80.71} & \textbf{78.55} & \textbf{69.10} \\
\cmidrule(r){2-2}\cmidrule(r){3-4}\cmidrule(l){5-6}\cmidrule(l){7-8}\cmidrule(l){9-10}\cmidrule(l){11-12} \cmidrule(l){13-14} \cmidrule(l){15-16}
&SeqTraL-FE  &   88.43          & 85.43          & 81.86          & 76.60          & 82.97          & 78.46          & 80.15          & 75.24          & 88.08          & 85.57          & 85.61          & 81.77          & 81.20          & 73.70          \\
&SeqTraL-UF     &88.50          & 85.36          & 82.52          & 76.79          & \textbf{83.83} & \textbf{79.24} & \textbf{80.79} & \textbf{75.65} & \textbf{88.87} & 86.30          & 85.78          & 81.54          & 81.51          & 73.65          \\
SeqTraL&SeqTraL-DL        & \textbf{89.06} & \textbf{85.88} & 82.57          & 76.66          & 83.36          & 78.57          & 80.29          & 74.89          & 88.78          & 86.14          & \textbf{86.25} & \textbf{81.85} & 81.17          & 73.10          \\
&SeqTraL-FT       &88.80          & 85.47          & \textbf{82.66} & \textbf{76.83} & 83.79          & 78.95          & 80.13          & 75.11          & 88.86          & \textbf{86.31} & 86.03          & 81.64          & \textbf{81.84} & \textbf{73.94} \\
\cmidrule(r){2-2}\cmidrule(r){3-4}\cmidrule(l){5-6}\cmidrule(l){7-8}\cmidrule(l){9-10}\cmidrule(l){11-12} \cmidrule(l){13-14} \cmidrule(l){15-16}
&MTL-Case    & \textbf{86.73} & \textbf{82.47} & \textbf{80.49} & \textbf{74.08} & \textbf{80.73} & \textbf{75.52} & -              & -              & 86.45          & 83.82          & 82.86          & 77.46          & 76.15          & 65.36          \\
Multi-tasking&MTL-Label     & 86.13          & 81.55          & 79.86          & 72.72          & 80.07          & 73.92          & 75.52          & 69.30          & \textbf{87.44} & \textbf{84.62} & 83.08          & 77.94          & 76.02          & 65.20          \\
&MTL-Morph      & 86.30          & 82.23          & 80.02          & 73.55          & 80.49          & 74.70          & \textbf{77.33} & \textbf{71.05} & 87.00          & 84.22          & \textbf{83.25} & \textbf{78.75} & \textbf{76.71} & \textbf{66.69} \\
\cmidrule(r){2-2}\cmidrule(r){3-4}\cmidrule(l){5-6}\cmidrule(l){7-8}\cmidrule(l){9-10}\cmidrule(l){11-12} \cmidrule(l){13-14} \cmidrule(l){15-16}
&BiAFF    &86.61          & 82.23          & 79.89          & 73.13          & 80.51          & 75.11          & 75.66          & 69.42          & 87.33          & 84.48          & 82.63          & 77.83          & 75.47          & 65.77          \\
&+Pretraining     & \textbf{93.61} & \textbf{90.85} & \textbf{86.04} & \textbf{79.55} & \textbf{89.05} & \textbf{83.80} & 79.88          & 74.65          & 88.79          & 86.02          & 85.85          & 81.66          & 81.63          & 73.86          \\
&+MTL      &89.99          & 86.49          & 82.47          & 76.24          & 84.35          & 79.74          & 80.33          & 75.15          & 88.42          & 85.94          & 85.91          & 81.56          & 81.30          & 73.49          \\
Prop. system&+SeqTraL      &90.31          & 86.70          & 82.70          & 76.57          & 84.58          & 80.15          & \textbf{80.79} & \textbf{75.65} & \textbf{88.87} & \textbf{86.30} & \textbf{86.05} & \textbf{81.85} & \textbf{81.84} & \textbf{73.94} \\
&+Self-training      & 89.83          & 86.09          & 82.08          & 75.92          & 84.12          & 79.66          & 80.08          & 75.24          & 88.78          & 86.07          & 85.73          & 81.77          & 79.89          & 72.28          \\
&+Data. aug. &89.11          & 85.87          & 82.08          & 75.92          & 84.12          & 79.66          & 79.56          & 73.53          & 88.31          & 84.67          & 85.73         & 81.77          & 79.52          & 71.89  \\
\cmidrule(r){2-2}\cmidrule(r){3-4}\cmidrule(l){5-6}\cmidrule(l){7-8}\cmidrule(l){9-10}\cmidrule(l){11-12} \cmidrule(l){13-14} \cmidrule(l){15-16}
Evaluation&BiAFF      & 87.10           & 83.06          & 80.92         & 75.02          & 80.31          &74.16 & 77.73          & 70.72          & 88.50          & 85.32          & 80.92          &75.02         & 79.33         & 67.92          \\
on test set&Prop. system & \textbf{93.66}           & \textbf{90.68}          & \textbf{86.43}         & \textbf{79.88}          & \textbf{88.50}          &\textbf{82.67} & \textbf{82.52}          & \textbf{77.07}          & \textbf{89.31} & \textbf{86.38}          & \textbf{87.50}          &\textbf{82.95}         & \textbf{83.59}         & \textbf{74.83}   \\ 
\hline
\end{tabular}}
    \caption{Evaluation of low-resource strategies on 7 languages. Experiments are first performed on dev set to find best performing combination of strategies for each language. The best results from strategies from each family are bold and statistically significant compared to its peer baselines belonging to the same family as per t-test ($p < 0.01$). The second last block shows ablations when the best variant from each family is added to the ensembled system. For example, +Data. aug. refers to the system with the best variant from all 5 strategies. The best performing system as per dev set is finally compared with BiAFF on the test set. XLM-R is not compatible with 3 languages and case information of Gothic (\textit{got}) language is missing; hence we do not report their results.}
    \label{table:multilingual_results}
    \end{small}
\end{table*}
\paragraph{Sequential Transfer Learning (SeqTraL):}
Following \newcite{sandhan-etal-2021-little}, we pretrain three encoders (similar to BiAFF) on three sequence labelling auxiliary tasks and integrate them with the BiAFF encoder using a gating mechanism. We adapt these pretrained encoders with various optimization schemes, proposed for reducing a catastrophic forgetting~\cite{french1999catastrophic,MCCLOSKEY1989109}.
\textbf{SeqTraL-FE:} We treat newly integrated layers as Feature Extractors (FE) by freezing them.  
 \textbf{SeqTraL-UF:} Gradually Unfreeze (UF) these new layers in the top to down order~\cite{howard-ruder-2018-universal,felbo-etal-2017-using}. \textbf{SeqTraL-DL:} The discriminative learning rate (DL)  is used for newly added layers~\cite{howard-ruder-2018-universal}, the learning rate is decreased from top-to-bottom layers. \textbf{SeqTraL-FT:} The default learning rate is used to fine-tune all newly added layers.

\paragraph{Cross/mono-lingual Pretraining:}
We experiment with two multilingual pretrained models, namely, the multilingual BERT \cite[\textbf{mBERT}]{devlin-etal-2019-bert} based system \cite{kondratyuk-straka-2019-75} and the XLM-Roberta \cite[\textbf{XLM-R}]{conneau-etal-2020-unsupervised} based system \cite{nguyen2021trankit}. 
We also consider supervised pretraining specially tailored for low-resource dependency parsing \cite[\textbf{LCM}]{sandhan-etal-2021-little} which essentially combines three sequence labelling auxiliary tasks. 
We pretrain it on 1,000 morphologically tagged data points without dependency annotations. 
\begin{figure*}[!tbh]
\centering
\subfloat[\label{fig:tagging_schem}]{\includegraphics[width=0.45\linewidth]{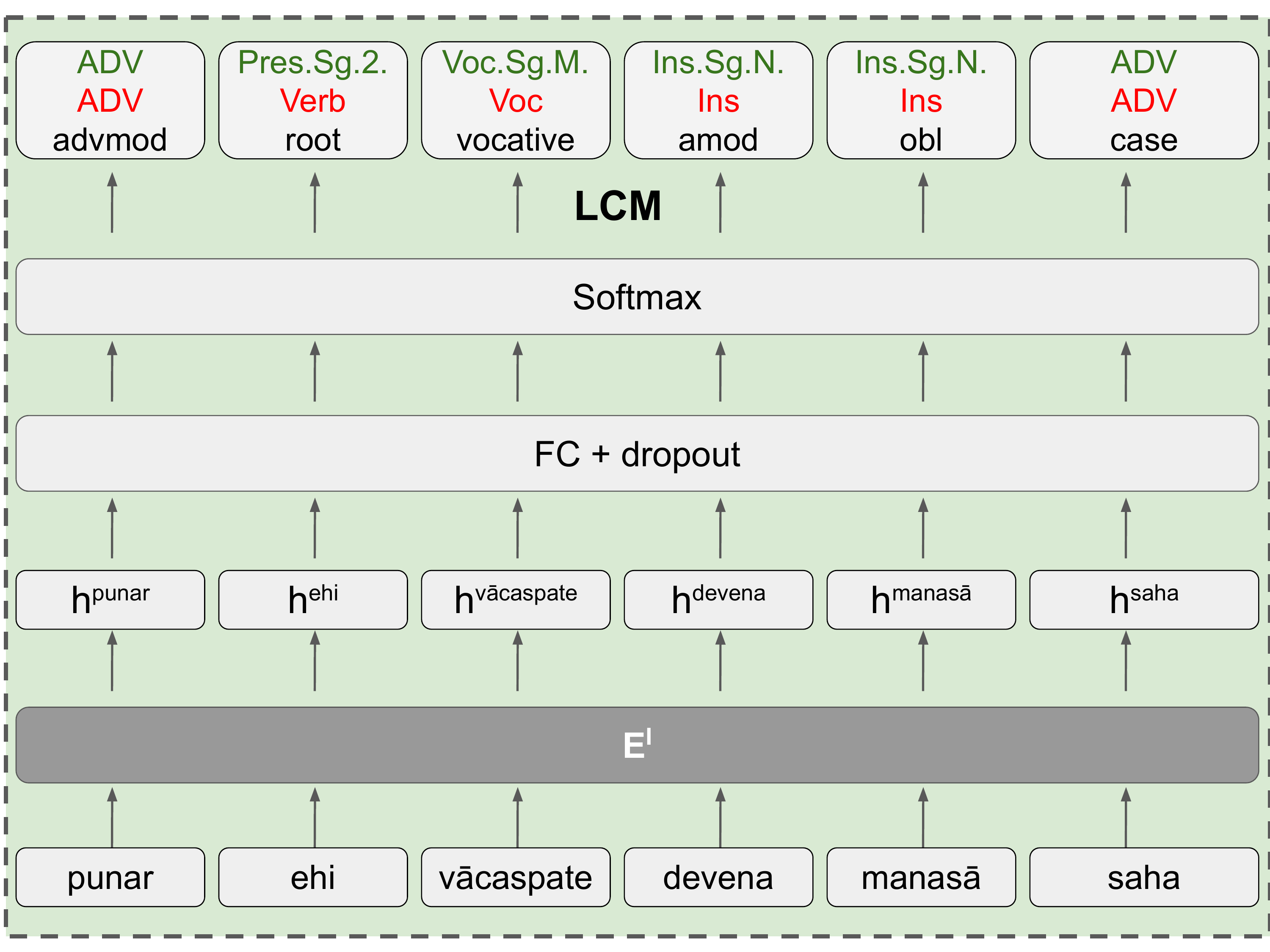}}
\subfloat[\label{fig:gating}]{\includegraphics[width=0.45\linewidth]{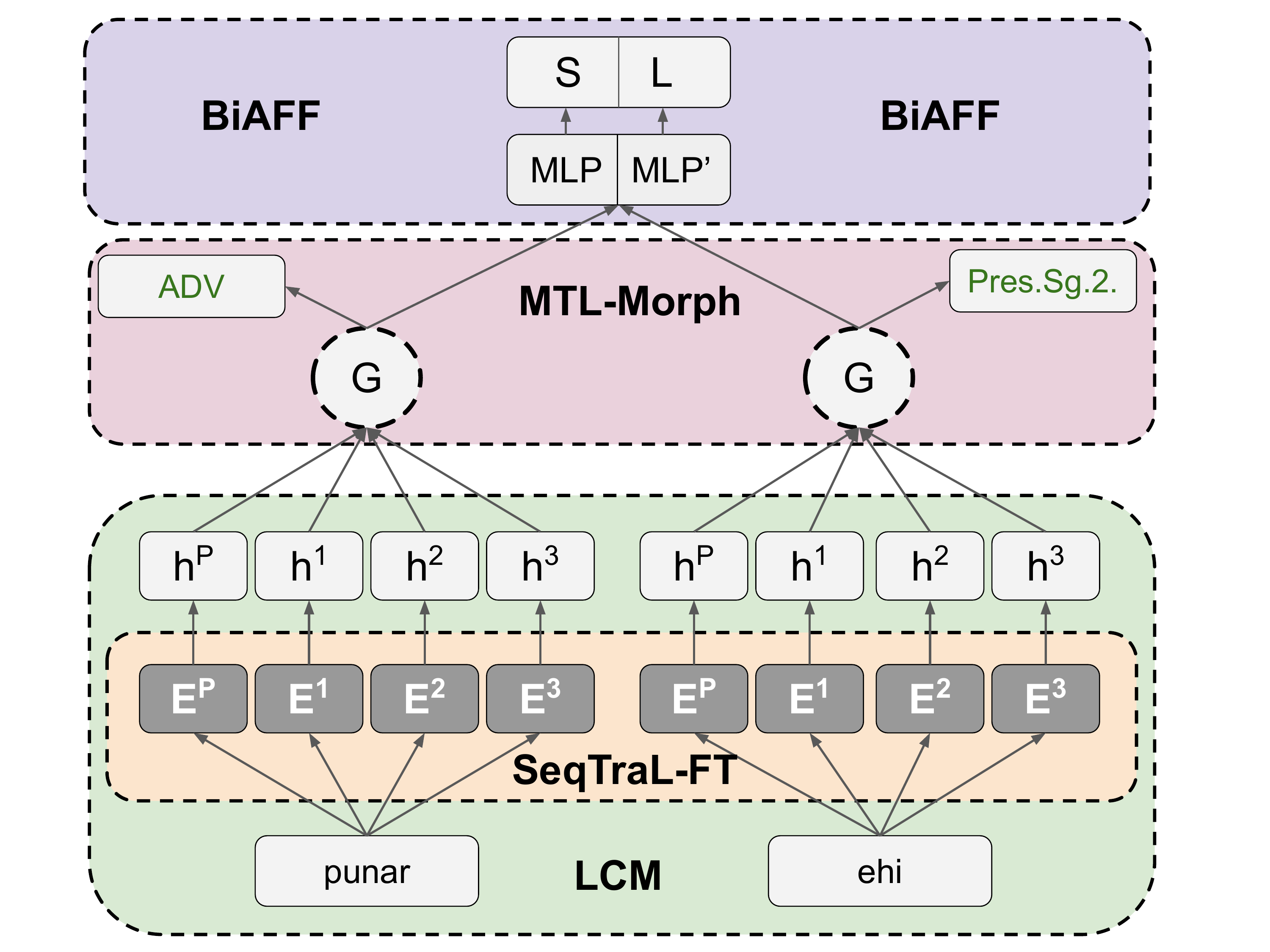}}
\caption{The ensembled system for Sanskrit. Translation: ``Oh V\={a}caspate! Come again with divine mind".
}
\label{fig:model}
\end{figure*}

\paragraph{Self-training:} 
Another line of modelling focuses on self-training~\cite{goldwasser-etal-2011-confidence,clark-etal-2018-semi,rybak2018semi} to overcome the bottleneck of task-specific labelled data. Earlier attempts failed to prove effectiveness of self-training for dependency parsing~\cite{rush-etal-2012-improved}. However, \newcite[\textbf{CVT}]{clark-etal-2018-semi} and \newcite[\textbf{DCST}]{rotman2019deep}, show successful application, thus, we consider these two systems. Also, we generate dependency data by applying a pretrained BiAFF system on 1000 unlabelled data points. We augment this predicted data with gold data and retrain BiAFF in \textbf{Self-Train} setting.

\paragraph{Multi-task Learning:}
We simultaneously train BiAFF and a sequence labelling based auxiliary task in a multi-task setting (\textbf{MTL}).  We experiment with the following auxiliary tasks: prediction of the morphological label (\textbf{MTL-Morph}),
dependency relation between a word and its head (\textbf{MTL-Label}) and the case label (\textbf{MTL-Case}).

\paragraph{Data Augmentation:}
\label{data_augmentation}
\newcite{sahin-steedman-2018-data} introduce \textbf{Cropping:} delete some parts of a sentence to create multiple short meaningful sentences, and \textbf{Rotation:} permute the siblings of headword restricted to a set of relations. 
Both operations modify a set of words or configurational information; however, they do not change the dependencies. 
\textbf{Nonce:} \newcite{gulordava-etal-2018-colorless} propose to create nonce sentences by substituting a few words which share the same syntactic labels. For each variant, we use additional 1,000 augmented data points. 

\noindent\paragraph{Results on multilingual experiments:}
 Table~\ref{table:multilingual_results} first reports results of all 5 strategies on dev set of 7 languages.
 Next, the second last block of Table~\ref{table:multilingual_results} (Prop. system) shows ablations on dev set where the best variant from each family is gradually added into the ensembled system. For example, \textit{+Data.aug.} row refers to the system with the best variant from all 5 strategies. Finally, the best performing system as per dev set is compared with BiAFF on the test set.
 We observe that (1) the best performing variant from augmentation, SeqTraL and MTL families is language dependent. (2) DCST variant of self-training wins over its peer for all the languages. (3) XLM-R outperforms for the languages which are covered in its pretraining (except Sanskrit\footnote{Maybe due to limited coverage of corpus for Sanskrit.}) and LCM outperforms for the rest of the languages which are truly low-resource. (4) Notably, we find effective generalization ability of the proposed approach on languages covered in cross-lingual pretraining (only pretraining helps)
 and for the rest of the languages (pretraining, MTL and SeqTraL helps). 

\section{Application on Sanskrit}
\label{experiments}
\noindent\textbf{Data:} 
We use two standard benchmark datasets available for Sanskrit. We use 1,700, 1,000 and 1,300 sentences (prose domain) from the Sanskrit Treebank Corpus \cite[STBC]{kulkarni2010designing} as train, dev and test set, respectively.
 We also evaluate on the Vedic Sanskrit Treebank \cite[VST]{hellwig-etal-2020-treebank} consisting of 1,500 , 1,024 and 1,473 sentences (poetry-prose mixed) as train, dev and test data, respectively. For both data, the final results on the test set are reported using systems trained with combined gold train and dev set.\\
   
\noindent\textbf{Baselines:}
We use \newcite[\textbf{YAP}]{more-etal-2019-joint} and \newcite[\textbf{L2S}]{chang2016} from transition-based dependency parsing family.
  \newcite[\textbf{BiAFF}]{DBLP:conf/iclr/DozatM17} is a graph-based approach with BiAFFINE attention mechanism.  \newcite[\textbf{MG-EBM}]{krishna-etal-2020-keep} extends \newcite[\textbf{Tree-EBM-F}]{amrith21} using multi-graph formulation. 
  Systems marked with (*) are hybrid systems which leverage linguistic rules from P\={a}\d{n}ini.\\
  
\noindent\textbf{The ensembled system:}
Figure~\ref{fig:model} shows the ensembled system for Sanskrit as per Table~\ref{table:multilingual_results}. It consists of two steps, namely, pretraining (\textbf{LCM} \crule[lightg]{0.25cm}{0.25cm}) and integration. As shown in Figure~\ref{fig:tagging_schem}, \textbf{LCM} pretrains three encoders $E^{(1)-(3)}$ using three independent auxiliary tasks, namely,  morphological label prediction, case label prediction and relation label prediction.  Thereafter, as shown in Figure \ref{fig:gating}, these pretrained encoders are integrated with the BiAFF  encoder $E^{(P)}$  using a gating mechanism  as employed in ~\newcite{sato-etal-2017-adversarial}. We use \textbf{SeqTraL-FT} \crule[lorg]{0.25cm}{0.25cm} optimization scheme to update the weights of these four encoders. Next, \textbf{MTL-Morph} \crule[lpink]{0.25cm}{0.25cm} component adds morphological tagging as an auxiliary task to inject complementary signal in the model. Finally, the combined representation of a pair of words in passed to \textbf{BiAFF} \crule[lvio]{0.25cm}{0.25cm} to calculate probability of arc score (S) and label (L).
 \begin{table}[h]
\centering
\resizebox{0.45\textwidth}{!}{
\begin{tabular}{|c|c|c|c|c|}
\hline
 &\multicolumn{2}{c|}{\textbf{STBC}}
&
\multicolumn{2}{c|}{\textbf{VST}} \\\hline
\textbf{System} &\textbf{UAS}&\textbf{LAS}    & \textbf{UAS} &\textbf{LAS}      \\\hline
YAP                & 75.31 & 66.02&70.37&56.09\\
L2S                & 81.97 & 74.14&72.44&62.76 \\
Tree-EBM-F                & 82.65 & 79.28&-&- \\
BiAFF              & 85.88 & 79.55&77.23&67.68 \\
Ours              & \textbf{88.67} & \textbf{83.47}&\textbf{79.71}&\textbf{69.89} \\
\hline
Tree-EBM-F*               &  \textit{85.32} & \textit{83.93} &-&- \\
MG-EBM* & \textit{87.46} & \textit{84.70} &-&- \\ \hline
\end{tabular}} 
    \caption{Results on test set for Sanskrit. \textit{Hybrid} systems, marked with (*) use extra-linguistic knowledge and are not directly comparable with our system. Our results are statistically significant compared to BiAFF as per t-test ($p < 0.01$). Results are averaged over 3 runs.} 
     \label{table:san_results}
\end{table}

\noindent\textbf{Results:}
On STBC, the ensembled system outperforms the state of the art \textit{purely data-driven} system (BiAFF) by 2.8/3.9 points (UAS/LAS) absolute gain.
Interestingly, it also supersedes the performance of the \textit{hybrid} state of the art system \cite[\textbf{MG-EBM}]{krishna-etal-2020-keep} by 1.2 points (UAS) absolute gain and shows comparable performance for LAS metric. We observe that performance of transition-based systems (\textbf{YAP/L2S}) is significantly low compared to graph-based counterparts (\textbf{BiAFF/Ours}). We also obtain a similar performance trend for VST data.
The VST data is a mixture of dependency labelled trees from both poetry and prose domain. As a result, the overall performance for VST is low compared to STBC due to loss of configurational information.\footnote{We do not evaluate Tree-EBM-F* and MG-EBM* on VST data due to the unavailability of the codebase.}

\section{Conclusion and Discussion}
We focused on low-resource dependency parsing for multiple languages. We found that our ensembled system can benefit the languages not covered in pretrained models. While multi-lingual pretraining (mBERT and XLM-R) is helpful for the languages covered in pretrained models, LCM pretraining (which simply uses an additional 1,000 morphologically tagged data points) is helpful for the remaining languages. Thus, these findings would help community to pick strategies suitable for their language of interest and come up with robust parsing solutions.
Specifically for Sanskrit, our ensembled system superseded the performance of the state-of-the-art \textit{hybrid} system MG-EBM* by 1.2 points (UAS) absolute gain and showed comparable performance in terms of LAS. 

\paragraph{Limitations:} We could not evaluate on complete UD due to limited available compute resources (single GPU), hence we selected 7 representative languages for our experiments.

\paragraph{Ethics Statement:}
We do not foresee any ethical concerns with the
work presented in this manuscript.

\section*{Acknowledgement}
We thank Amba Kulkarni for providing Sanskrit dependency treebank data, Anupama Ryali for \`{S}i\`{s}up\={a}lavadha dataset.  We are grateful to Amrith Krishna (Uniphore) for helping us with the initial discussions on this work. We thank Tushar Sandhan (IIT Kanpur), Narein Rao (IIT Kanpur), Rathin Singha (UCLA)  and the anonymous reviewers for their constructive feedback towards improving this work. The work of the first author is supported by the TCS Fellowship under the Project TCS/EE/2011191P. The work was supported in part by the National Language Translation Mission (NLTM): Bhashini project by Government of India.

\bibliography{anthology,custom}
\bibliographystyle{acl_natbib}
\appendix

\end{document}